\documentclass[review]{elsarticle}

\newcommand{\eg}{e.g.\@}
\newcommand{\ie}{i.e.\@}
\newcommand{\etal}{et al.\@}

\usepackage{amsmath,amsfonts,bm}

\def\eqref#1{equation~\ref{#1}}

\def\1{\bm{1}}

\DeclareMathAlphabet{\mathsfit}{\encodingdefault}{\sfdefault}{m}{sl}
\SetMathAlphabet{\mathsfit}{bold}{\encodingdefault}{\sfdefault}{bx}{n}

\usepackage{amsmath, amsthm}
\usepackage{setspace}
\usepackage{amssymb}
\usepackage{color}
\usepackage{rotating}
\usepackage{placeins}
\usepackage{multirow}
\usepackage{lineno}
\usepackage{graphicx}
\usepackage{color,xcolor}
\usepackage{bm}
\usepackage{url}
\usepackage{epsfig}
\usepackage{threeparttable}
\usepackage{booktabs}
\usepackage{amsthm}
\usepackage{thmtools}
\usepackage{booktabs} 
\usepackage{paralist}
\usepackage{diagbox}
\usepackage{subfigure}

\usepackage[linesnumbered,ruled,vlined]{algorithm2e}
\SetKwInput{KwInput}{Input}              
\SetKwInput{KwOutput}{Output}

\journal{Pattern Recognition}

\begin{document}

\begin{frontmatter}

\title{Attentional Prototype Inference for Few-Shot Segmentation}

\author[1]{Haoliang Sun}

\author[1]{Xiankai Lu}

\author[2]{Haochen Wang}

\author[1]{Yilong Yin\corref{cor1}}

\author[3,4,5]{Xiantong Zhen}

\author[3]{Cees~G.~M.~Snoek}

\author[4]{Ling Shao}


\cortext[cor1]{Corresponding author.}

\address[1]{School of Software, Shandong University, Jinan, China}
\address[2]{Alibaba Group, Beijing, China}
\address[3]{Informatics Institute, University of Amsterdam, The Netherlands}
\address[4]{Inception Institute of Artificial Intelligence, U.A.E.}
\address[5]{United Imaging Intelligence, Beijing, China}

\begin{abstract}\label{sec:abstract}
This paper aims to address few-shot segmentation. While existing prototype-based methods have achieved considerable success, they suffer from uncertainty and ambiguity caused by limited labeled examples. In this work, we propose attentional prototype inference (API), a probabilistic latent variable framework for few-shot segmentation. We define a global latent variable to represent the prototype of each object category, which we model as a probabilistic distribution. The probabilistic modeling of the prototype enhances the model's generalization ability by handling the inherent uncertainty caused by limited data and intra-class variations of objects. To further enhance the model, we introduce a local latent variable to represent the attention map of each query image, which enables the model to attend to foreground objects while suppressing the background. The optimization of the proposed model is formulated as a variational Bayesian inference problem, which is established by amortized inference networks. We conduct extensive experiments on four benchmarks, where our proposal obtains at least competitive and often better performance than state-of-the-art prototype-based methods. We also provide comprehensive analyses and ablation studies to gain insight into the effectiveness of our method for few-shot segmentation.

\end{abstract}
\begin{keyword}
	Few-Shot Segmentation, Variational Inference, Probabilistic Model, Latent Attention
\end{keyword}

\end{frontmatter}

\section{Introduction}\label{sec:introduction}
Semantic segmentation \cite{brostow2009semantic} has been a fundamental problem in computer vision with widespread application potential in a great variety of areas, e.g., autonomous driving and scene understanding. Existing models based on deep convolutional neural networks and trained on massive amounts of manually labeled images, e.g., \cite{chen2017deeplab,long2015fully}, have obtained impressive results. However, since their performance relies heavily on access to a large number of pixel-wise annotations, it remains challenging to achieve desirable results in practice when the training data is scarce. Therefore, few-shot segmentation 
\cite{fan2022self} has emerged as a popular task to address the annotation scarcity issue of traditional semantic segmentation methods. 

Few-shot segmentation generalizes the idea of few-shot classification under the setting of meta-learning \cite{luo2022meta}. In meta-learning, the dataset is split into meta-training, meta-validation, and meta-testing sets. Few-shot segmentation solutions usually sample an episode consisting of a support and query set from these meta-sets to train a learning procedure that takes the support set as input and produces the prediction on the query set. The model can then achieve effective adaption to new tasks at test time. In this paper, we focus on few-shot segmentation, where we aim to segment the object of an unseen category in a query image with the support of only a few annotated images.

\begin{figure}[]
	\centering
	\includegraphics[width=0.69\linewidth]{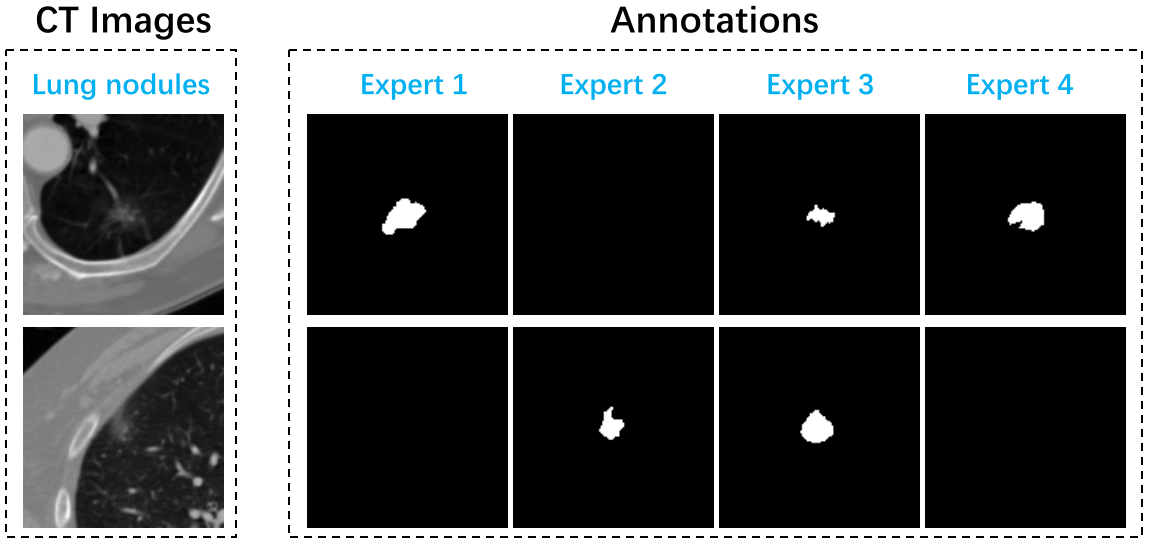}
	\caption{Intrinsic ambiguities exist in images as four experts give diverse lesion annotations.}
	\label{fig:example}
\end{figure}

To alleviate the scarcity of annotated data, most existing works extract supervisory information for the objects from a small set of support images. Inspired by the prototype theory from cognitive science \cite{rosch1973natural} and prototype networks for few-shot classification \cite{snell2017prototypical}, many segmentation models are designed to learn a prototype vector to represent a category in the support set
~\cite{li2021adaptive}. The optimization goal is then to obtain a shared feature extractor that generalizes to the segmentation of new objects \cite{dong2018few_pl}. 
While prototype-based methods have shown great efficiency in few-shot segmentation, there still exist three major deficiencies. 1) existing methods map the support images into a deterministic prototype vector, which is often ambiguous and vulnerable to noise under the few-shot setting, especially for one-shot learning tasks. As illustrated in Figure \ref{fig:example}, intrinsic ambiguities usually exist in images. Deterministic models neglect those ambiguities and merely provide the most likely hypothesis that might cause sub-optimal decision. 2) The prototype vector loses the structure information of the object in the query image. 3) A deterministic prototype vector contains only its first-order statistics, which are unable to represent large intra-class variations of objects in the same category. To address these issues, we propose a probabilistic latent variable framework, referred to as attentional prototype inference (API), for few-shot segmentation.

%
%

We make three contributions in this work. (1) We provide a fully probabilistic framework for few-shot segmentation. We introduce a global latent variable into this model, which represents the prototype of each object category. The probabilistic framework models the prototype as a distribution rather than a vector, making it more robust to noise and better equipped to deal with ambiguity than the deterministic model. The probabilistic prototype also better represents object categories by effectively capturing intra-class variations. 
(2) The second contribution is that we introduce a variational attention mechanism. We define the attention vector as the local latent variable associated with each image and infer its probabilistic distribution, which is jointly estimated within the same framework as the variational prototype inference. The main motivation of variational attention is to enable the model to capture the appearance variation of an object. (3) Our third contribution is to formulate the optimization as a variational inference problem to jointly estimate posteriors over latent variables. The optimization objective is built upon a newly derived evidence lower bound (ELBO), which fits the few-shot segmentation problem well and offers a principled way to model prototypes and attention for few-shot segmentation.

To evaluate our attentional prototype inference, we conduct extensive experiments on four benchmarks, i.e., Pascal-$5^i$, MS-COCO, FSS-1000, and LIDC-IDRI. The comparison results show that our attentional prototype inference achieves at least competitive and often better performance than state-of-the-art prototype-based methods on both the 1-shot and 5-shot segmentation tasks, demonstrating its effectiveness for few-shot segmentation. Quantitative comparison in terms of the cross energy distance on the medical image dataset shows the good statistical property of the proposed algorithm, which can faithfully handle the uncertainty stemming from intrinsic ambiguities in images. We also conduct ablation studies to gain insight into the proposed attentional prototype inference by demonstrating the benefit of different model components to the overall performance.



\section{Related Work}\label{sec:related}

\subsection{Many-Shot Semantic Segmentation} 
Semantic segmentation aims to segment a given image into several pre-defined classes and is often regarded as a pixel-level classification task~\cite{brostow2009semantic}. State-of-the-art semantic segmentation methods \cite{ZHOU2022108290,lu2021segmenting} based on deep convolutional neural networks have achieved astonishing success. The fully convolutional network (FCN)~\cite{long2015fully} was the first model to introduce end-to-end convolutional neural networks into segmentation tasks. The essential innovation in FCN is replacing the fully-connected layer with a fully convolutional architecture to preserve the spatial information for better performance. Follow-up efforts have attempted to aggregate multiple pixels to explicitly model context. For example, DeepLab \cite{chen2017deeplab} introduces a dilated convolution operation to enlarge the perception field while maintaining the resolution, and PSPNet~\cite{zhao2017pyramid} employs a pyramid pooling module to aggregate multi-scale context information. Another novel pyramid module~\cite{zhang2021gpnet} is designed to capture and filter the multi-scale information in a gated and pair-wise manner. 


Though they achieve impressive performance, these methods heavily rely on labeled training samples with pixel-level annotations. However, pixel-level annotations are expensive and difficult to obtain. Moreover, the deep semantic segmentation models usually perform modestly on new categories of objects that are unseen in the training set, which restricts their use in practical applications. 

\subsection{Few-Shot Segmentation}
%
%
Few-shot segmentation aims to segment images from arbitrary classes by learning transferable knowledge from scarce annotated support images, which has recently gained popularity in computer vision applications.
Shaban \etal~\cite{shaban2017one_bmvc} introduced the first few-shot segmentation network based on a two-branch architecture, which uses a support branch to predict the parameters of the last layer of the query branch for segmentation. Recent works \cite{siam2019amp_amp} also follow this two-branch architecture for few-shot segmentation. Dong \etal~\cite{dong2018few_pl} generalized the idea of prototype networks~\cite{snell2017prototypical} from few-shot recognition for few-shot segmentation. They designed the PLNet, in which the first branch learns a prototype vector that takes images and annotations as input and outputs the prototype. Meanwhile, the second branch takes both a new image and the prototype as input and outputs the segmentation mask. Since then, prototype-based methods have been further developed using different strategies 
\cite{li2021adaptive, okazawa2022interclass}.

To achieve sufficient representation for the class prototype, Yang et al.~\cite{yang2020prototype} designed prototype mixture models to correlate diverse image regions with multiple prototypes. 
Tian et al.~\cite{tian2020pfenet} proposed the feature enrichment module to overcome spatial inconsistency. 
Li et al. \cite{LI2021107882} designed a prototype alignment regularization to generate a more consistent prototype between support and query.
Okazawa \cite{okazawa2022interclass} proposed to reduce the similarity between prototypes of each class and leverage the relationship between classes in a batch.
These works have demonstrated the effectiveness of prototype learning for few-shot segmentation. However, a deterministic prototype vector is not sufficiently representative for capturing the categorical concepts of objects and therefore can cause bias and reduced generalization when objects in the same categories vary. 
In this work, we develop a variational attention mechanism by placing a distribution over the attention vector, which enables the model to better capture the appearance variation of individual objects. 

Other strategies have been adopted in recent years and achieve substantial improvement in this task. Singh et al.~\cite{singh2021metamed} adopted a gradient-based meta-learning algorithm and integrated augmentations for few-shot medical imaging segmentation. Min et al. \cite{Min_2021_ICCV} introduced hypercorrelation squeeze networks with 4D convolution layers to characterize correspondences in multiple visual aspects between support and query images. Recently, Swin Transformer~\cite{hong2022cost} has been applied to handle a high-dimensional correlation map and achieved superior performance on those datasets. Gaussian process~\cite{johnander2022dense} has also been introduced to extract detailed relations among support images. Fan \etal~\cite{fan2022self} designed a novel self-support matching strategy to resolve the issue of Gestalt principle.

\subsection{Variational Inference}

Variational auto-encoder (VAE) \cite{kingma2013auto} is a generative model that introduces variational inference (VI)  \cite{ lim2021variational} into the learning of directed graphical models. It has also been broadly applied in segmentation tasks. For example, Kohl \etal~proposed the probabilistic U-net~\cite{kohl2018probabilistic} which combines C-VAE with U-Net for image segmentation. It learns a distribution over the segmentation masks to handle ambiguities, especially for medical images. 
Zhang \etal~\cite{zhang2019variational} deployed a latent variable to denote the distribution of the entire dataset, which is inferred from the support set.
They also showed that their variational learning strategy can be modified to classify proposals for instance segmentation.

We address few-shot segmentation based on prototypes using a probabilistic latent variable model. We treat the prototype that represents the concept of the object category as a global latent variable, which is modeled as a distribution instead of a single deterministic vector. We further introduce a local latent variable to generate the attention map, which is learned for each image to highlight foreground objects. We solve the whole model by variational Bayesian inference, in which latent variables are jointly learned in the same framework.

\section{Methodology}\label{sec:method}


We adopt the meta-learning setting to conduct few-shot segmentation. We learn a segmentation model on the meta-training set $\mathcal{D}_{\text{train}}$ and then evaluate it on the meta-testing set $\mathcal{D}_{\text{test}}$. Different from the traditional semantic segmentation, there is no overlap between the object categories in $\mathcal{D}_{\text{train}}$ and $\mathcal{D}_{\text{test}}$. To achieve few-shot segmentation, we follow the episodic paradigm for training and testing under a $k$-shot setting, where $k$ denotes the number of training images in an episode. 
In practice, we sample one episode each time from $\mathcal{D}_{\text{train}}$ for training or $\mathcal{D}_{\text{test}}$ for evaluation. 
Each episode is composed of a support set $\mathcal{S} = \{(\mathbf{x}_s^i,\mathbf{y}_s^i)\}_{i=1}^{k}$ and a query set $\mathcal{Q} = \{(\mathbf{x}_q,\mathbf{y}_q)\}$. Here, $\mathbf{x}_s^i \in \mathbb{R}^{h \times w \times 3}$ denotes the support image with a height of $h$ and width of $w$. $\mathbf{y}_s^i \in \mathbb{R}^{h \times w}$ is its corresponding support mask. Similarly, $\mathbf{x}_q$ is the query image and $\mathbf{y}_q$ is the associated ground-truth mask of the object to be segmented. The goal of the few-shot segmentation model is to extract transferable knowledge from the support set $\mathcal{S}$ and apply it to the segmentation of a query image $\mathbf{x}_q$. The predicted segmentation map is denoted as $\tilde{\mathbf{y}}_q$ for $\mathbf{x}_q$. 


\subsection{Attentional Prototype Inference}

From a probabilistic perspective, the purpose of few-shot segmentation is to find the prediction $\tilde{\mathbf{y}}_q $ that can maximize the probability of the conditional predictive distribution $p(\mathbf{y}_q|\mathbf{x}_q,\mathcal{S})$ over the segmentation map $\mathbf{y}_q$ for a given query image $\mathbf{x}_q$, when provided the support set $\mathcal{S}$.

\subsubsection{Probabilistic Modeling}
We introduce a latent variable  $\mathbf{z}$ to represent the class prototype, which is conditioned on the corresponding support set. By incorporating the latent variable, we have the conditional predictive  log-likelihood as follows
\begin{equation}
\begin{aligned}
\log p(\mathbf{y}_q|\mathbf{x}_q,\mathcal{S}) = \log\int p(\mathbf{y}_q|\mathbf{z},\mathbf{x}_q)p(\mathbf{z}|\mathcal{S})d\mathbf{z},
\end{aligned}
\label{logl}
\end{equation}
where $p(\mathbf{z}|\mathcal{S})$ is a conditional prior. The model in (\ref{logl}) provides a probabilistic modeling of prototypes for semantic segmentation, which was introduced in our preliminary work  \cite{wang2021variational}. In this way, our prototype serves as a global representation of an object category while previous ones do not take into account the local spatial structure of the image~\cite{dong2018few_pl}. 

To further enhance the model in (\ref{logl}), we introduce a local latent variable $\mathbf{m}_q$ to represent the attention map associated with each image, which highlights the foreground object. The conditional predictive log-likelihood with respect to the two latent variables takes the following form
\begin{equation}
\begin{aligned}
&\log p(\mathbf{y}_q|\mathbf{x}_q,\mathcal{S}) \\
&\quad= \log\int\!\!\!\!\int_{}^{}p(\mathbf{y}_q|\mathbf{z},\mathbf{m}_q,\mathbf{x}_q)p(\mathbf{m}_q|\mathbf{x}_q)p(\mathbf{z}|\mathcal{S})d\mathbf{z}d\mathbf{m}_q,
\end{aligned}
\label{logll}
\end{equation}
where we also deploy a conditional prior $p(\mathbf{m}_q|\mathbf{x}_q)$ for $\mathbf{m}_q$, since the attention maps should be specific for each individual query image $\mathbf{x}_q$.

However, these posteriors are intractable in practice. Thus, we introduce the variational posterior to approximate the true posteriors by minimizing their Kullback-Leibler (KL) divergence. We employ the variational distributions $q_{\phi_1}(\mathbf{z}|\mathbf{x}_q,\mathbf{y}_q, \mathcal{S})$ and $q_{\phi_2}(\mathbf{m}_q|\mathbf{x}_q,\mathbf{y}_q)$ for the prototype $\mathbf{z}$ and the attention map $\mathbf{m}_q$, respectively. 
%
%
By incorporating the variational posteriors into the conditional log-likelihood $\log p(\mathbf{y}_q|\mathbf{x}_q,S)$ of (\ref{logll}), we arrive at
\begin{equation}
\begin{aligned}
&\log p(\mathbf{y}_q|\mathbf{x}_q,S) \\
&\quad= \log\int\!\!\!\!\int p(\mathbf{y}_q|\mathbf{z},\mathbf{m}_q,\mathbf{x}_q)p(\mathbf{m}_q|\mathbf{x}_q)p(\mathbf{z}|\mathcal{S}) \\ 
&\quad\quad\quad\quad\quad\quad 
\frac{q_{\phi_1}(\mathbf{z}|\mathbf{x}_q,\mathbf{y}_q, \mathcal{S}) 
q_{\phi_2}(\mathbf{m}_q|\mathbf{x}_q,\mathbf{y}_q)}
{q_{\phi_1}(\mathbf{z}|\mathbf{x}_q,\mathbf{y}_q, \mathcal{S})
q_{\phi_2}(\mathbf{m}_q|\mathbf{x}_q,\mathbf{y}_q)}   \;        d\mathbf{z}d\mathbf{m}_q.
\end{aligned}
\end{equation}
Applying Jensen's inequality gives rise to the ELBO as follows
\begin{equation}
\begin{aligned}
&\log p(\mathbf{y}_q|\mathbf{x}_q,S) \\
 &\quad \geq \int q_{\phi_2}(\mathbf{m}_q|\mathbf{x}_q,\mathbf{y}_q)   \int
 q_{\phi_1}(\mathbf{z}|\mathbf{x}_q,\mathbf{y}_q, \mathcal{S})  \\ 
&\quad\quad\quad\quad   \frac{\log \big[p(\mathbf{y}_q|\mathbf{z},\mathbf{m}_q,\mathbf{x}_q)\big] p(\mathbf{z}|\mathcal{S}) p(\mathbf{m}_q|\mathbf{x}_q)}{q_{\phi_1}(\mathbf{z}|\mathbf{x}_q,\mathbf{y}_q, \mathcal{S})
q_{\phi_2}(\mathbf{m}_q|\mathbf{x}_q,\mathbf{y}_q)}    \;      d\mathbf{z}d\mathbf{m}_q&\\
& \quad = E_{q_{\phi_1}(\mathbf{z}|\mathbf{x}_q,\mathbf{y}_q, \mathcal{S}),q_{\phi_2}(\mathbf{m}_q|\mathbf{x}_q,\mathbf{y}_q)}\big[\log p(\mathbf{y}_q|\mathbf{z},\mathbf{m}_q,\mathbf{x}_q)\big] &\\
&\quad -D_{\mathrm{KL}}[q_{\phi_1}(\mathbf{z}|\mathbf{x}_q,\mathbf{y}_q, \mathcal{S})||p(\mathbf{z}|\mathcal{S})]\\ &\quad-D_{\mathrm{KL}}[q_{\phi_2}(\mathbf{m}_q|\mathbf{x}_q,\mathbf{y}_q)||p(\mathbf{m}_q|\mathbf{x}_q)]. 
\end{aligned}
\label{equa:1}
\end{equation}
The first term of the ELBO is the expectation of the log-likelihood of the conditional generative distribution $p(\mathbf{y}_q|\mathbf{z},\mathbf{m}_q,\mathbf{x}_q)$ based on the inferred prototypes $\bf{z}$ and attention maps $\mathbf{m}_q$. The second term is the KL divergence between the estimated posterior distribution $q_{\phi_1}(\mathbf{z}|\mathbf{x}_q,\mathbf{y}_q,\mathcal{S})$ and the prior distribution $p(\mathbf{z}|\mathcal{S})$. Minimizing this term encourages the model to leverage the object information for the segmentation of the query image. Minimizing the third term of the KL divergence enables the model to generate attention maps that highlight the foreground object. 
We derive the optimization objective based on the ELBO. 

\subsubsection{Optimization Objective}
Maximizing the ELBO can yield accurate predictions for the segmentation masks and narrow the gap between the posterior and the prior distributions. This encourages 1) the inferred prototype from the support dataset to match the full dataset by minimizing the first KL term; and 2) the inferred map from the query set to approach the one based merely on the query image by minimizing the second KL term. Based on the ELBO, we define the empirical objective function for optimization.

Given a batch of episodes, the empirical objective for stochastic optimization with the Monte Carlo estimation of expectations is as follows
\begin{equation}
\begin{aligned}
\mathcal{L} = \sum_i &\Big[ \frac{1}{LM} \sum\limits_{l=1}^{L}  \sum\limits_{j=1}^{M}  -\log p_{\psi}(\mathbf{y}^{(i)}_q|\mathbf{z}^{(l)},\mathbf{m}_q^{(j)},\mathbf{x}^{(i)}_q) &\\
&\quad +D_{\mathrm{KL}}[q_{\phi_1}(\mathbf{z}|\mathbf{x}^{(i)}_q,\mathbf{y}^{(i)}_q, \mathcal{S}^{(i)})||p_{\theta_1}(\mathbf{z}|\mathcal{S}^{(i)})]\\ &\quad +D_{\mathrm{KL}}[q_{\phi_2}(\mathbf{m}_q|\mathbf{x}^{(i)}_q,\mathbf{y}^{(i)}_q)||p_{\theta_2}(\mathbf{m}_q|\mathbf{x}^{(i)}_q)]  \Big],
\end{aligned}
\label{equa:loss}
\end{equation}
where $i$ indexes over the sampled episode in the meta-training set $\mathcal{D}_{\text{train}}$, $\mathbf{z}^{(l)} \sim q_{\phi_1}(\mathbf{z}|\mathbf{x}^{(i)}_q,\mathbf{y}^{(i)}_q, \mathcal{S}^{(i)})$ and $\mathbf{m}_q^{(j)} \sim q_{\phi_2}(\mathbf{m}_q|\mathbf{x}^i_q,\mathbf{y}^i_q)$ are the variables sampled from their variational distribution. $L, M$ are the number of samples. 
Generally, the parameters of a neural network are optimized jointly with stochastic gradient descent. However, it is usually intractable to calculate the gradient of the sampling operation. 
Therefore, we deploy the reparameterization trick \cite{kingma2013auto} to handle the non-differentiable problem of the sampling process. Specifically,
supposing the two variational posterior distributions take the form of a multivariate Gaussian with a diagonal covariance, the sampling process is formulated as
\begin{equation}\label{sampling}
\begin{aligned}
\left\{ \begin{array}{l}
    q_{\phi_1}(\mathbf{z}|\mathbf{x}_q,\mathbf{y}_q, \mathcal{S}) = \mathcal{N}(\mathbf{z}; \boldsymbol{\mu}_{\phi_1},\boldsymbol{\sigma}_{\phi_1}^2)\\
    \\
    q_{\phi_2}(\mathbf{m}_q|\mathbf{x}_q,\mathbf{y}_q) = \mathcal{N}(\mathbf{m}_q; \boldsymbol{\mu}_{\phi_2},\boldsymbol{\sigma}_{\phi_2}^2)\\
    \\
    p_{\theta_1}(\mathbf{z}|\mathcal{S}) = \mathcal{N}(\mathbf{z}; \boldsymbol{\mu}_{\theta_1},\boldsymbol{\sigma}_{\theta_1}^2) \\
    \\
    p_{\theta_2}(\mathbf{m}_q|\mathbf{x}_q) = \mathcal{N}(\mathbf{m}_q; \boldsymbol{\mu}_{\theta_2},\boldsymbol{\sigma}_{\theta_2}^2). \\
  \end{array} \right.
  \end{aligned}
\end{equation}
During training, the samples of the class prototype $\mathbf{z}$ are obtained by:
\begin{equation}\label{repa}
\mathbf{z}^{(l)} = \boldsymbol{\epsilon}^{(l)}\odot \boldsymbol{\sigma}_{\phi_1} + \boldsymbol{\mu}_{\phi_1},
\end{equation}
where $\odot$ denotes an element-wise multiplication and $\boldsymbol{\epsilon}^{(l)} \sim N(\boldsymbol{\epsilon}; \boldsymbol{0}, \boldsymbol{1})$. The same operation is also deployed for sampling $\mathbf{m}_q^{(j)}$. 

The first term of the empirical loss  $\mathcal{L}$ in (\ref{equa:loss}) is implemented as a least square loss in \cite{kingma2013auto}; we generalize this to a pixel-wise cross-entropy loss to penalize the difference between the predicted segmentation map $\tilde{\mathbf{y}}_q$ and the ground truth $\mathbf{y}_q$. The number of samples $L$ and $M$ are set to 1 during training to speed up the learning process. Since the KL terms minimize the discrepancy between two distributions, the prior networks can mimic the behavior of the posterior networks that produce effective prototypes or attention maps at training time. 


\subsubsection{Segmentation Map Inference}
The inference of segmentation maps varies between the learning and inference stages. At test time, instead of sampling from the variational posterior distributions, we draw $L$ samples of prototypes $\{\mathbf{z}^{(l)}\}_{l=1}^{L}$ from the prior $p_{\theta_1}(\mathbf{z}|\mathcal{S})$ and $M$ samples of attention vectors from $p(\mathbf{m}_q|\mathbf{x}_q)$. $\tilde{\mathbf{y}}_q$ is obtained by taking the average of $LM$ segmentation maps from these samples
\begin{equation}
\begin{aligned}
&\tilde{\mathbf{y}}_q = \frac{1}{LM}\sum\limits_{l=1}^{L}  \sum\limits_{j=1}^{M}  p_{\psi}(\mathbf{y}_q|\mathbf{x}_q,\mathbf{z}^{(l)},\mathbf{m}_q^{(j)}),\\ 
\end{aligned}
\label{equa:4}
\end{equation}
where 
\begin{equation}
    \mathbf{z}^{(l)}\sim p_{\theta_1}(\mathbf{z}|\mathcal{S})
\end{equation}
and
\begin{equation}
\mathbf{m}_q^{(j)}\sim p_{\theta_2}(\mathbf{m}_q|\mathbf{x}_q).
\end{equation}

\subsection{Neural Networks Implementation}
\label{sec:network}

\begin{figure*}[]
	\centering
	\includegraphics[width=.99\linewidth]{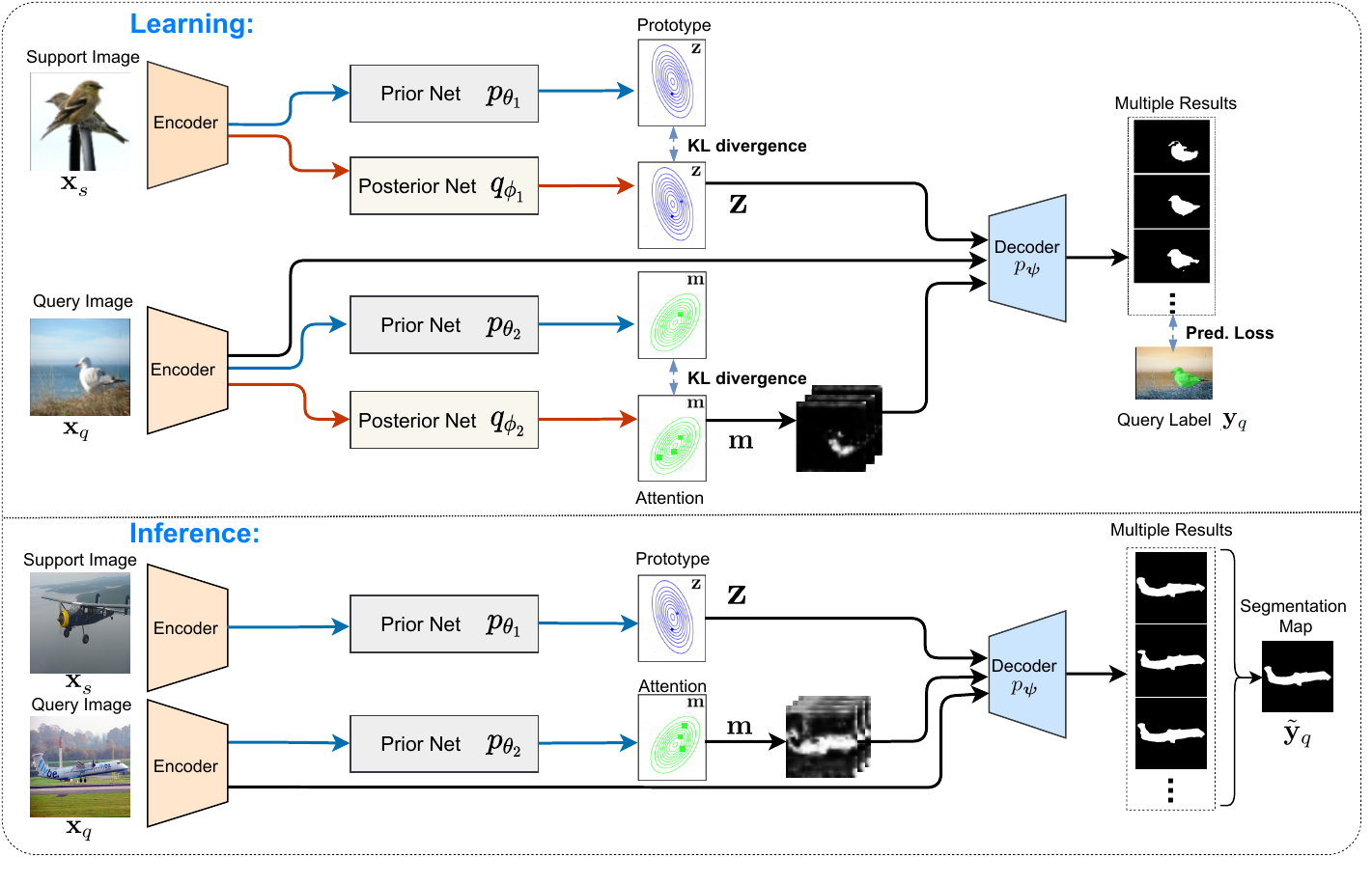}
	\vspace{-10pt}
	\caption{Attentional prototype inference for few-shot segmentation, implemented as the amortized neural network with an auto-encoder architecture. The prior nets and the posterior nets map the input images into the latent space obtaining the prior distributions $\{p_{\boldsymbol{\theta}}(\mathbf{z}|S), p_{\theta_2}(\mathbf{m}_q|\mathbf{x}^{(i)}_q)\}$ and the posteriors $\{q_{\phi_1}(\mathbf{z}|\mathbf{x}_q,\mathbf{y}_q, \mathcal{S}), q_{\phi_2}(\mathbf{m}_q|\mathbf{x}_q,\mathbf{y}_q)\}$. The segmentation net takes the query image $\mathbf{x}_q$, the sampled prototype vector $\mathbf{z}$ and the attention map $\mathbf{m}$ to generate a distribution of the segmentation map: $p_{\boldsymbol{\psi}}(\mathbf{y}_q|\mathbf{z},\mathbf{m},\mathbf{x}_q)$. By minimizing the predictive loss and the KL divergence between the prior and posterior, the whole model can be jointly learned in an end-to-end manner. During the inference stage, we obtain a set of samples of the prototype and attention map from the priors. Monte Carlo estimation is employed to predict an accurate segmentation map by averaging multiple outputs.}
	\label{fig:network}
\end{figure*}

We implement our attentional prototype inference with neural networks using the amortization technique \cite{kingma2013auto}, which is seamlessly integrated into the autoencoder architecture, as shown in Figure~\ref{fig:network}. We parameterize the distributions as factorized Gaussian distributions with diagonal covariance matrices. 

\textbf{Prior Networks.}
The prior network for prototypes embeds the support set $S$ into a function space, where the conditional prior distribution $p(\mathbf{z}|\mathcal{S})$ lies. The prior network for attention maps is encouraged to generate an effective attention map as $q_{\phi_2}(\mathbf{m}_q|\mathbf{x}_q,\mathbf{y}_q)$.
For the prior network of prototypes, we construct a multi-layer perceptron (MLP) $f_{\theta_1}(\cdot)$ with three fully connected layers. The extracted deep features of images are selected with their segmentation masks to obtain the foreground features. A permutation-invariant pooling layer~\cite{siam2019amp_amp} then squeezes them into a single vector $\bm{c}$. In this work, we assume that the prior follows a diagonal covariance Gaussian distribution. Given the single feature vector, the mean $\boldsymbol{\mu}_{\theta_1}$ and variance $\boldsymbol{\sigma}_{\theta_1}^2$ \textit{w.r.t.} $\mathbf{z}$ come from the output of the MLP:
\begin{equation}
\boldsymbol{\mu}_{\theta_1}, \boldsymbol{\sigma}_{\theta_1}^2 = f_{\theta_1}(\bm{c}).
\end{equation}

The main spirit behind the prior network of the attention maps is similar but we employ a transformer architecture \cite{bhat2020learning} to extract the structure information with high fidelity. This prior transformer contains a pixel-level self-attention and aggregates all pixels into a vector. Then a two-layer perceptron is concatenated to the output $\boldsymbol{\mu}_{\theta_2}$ and $\boldsymbol{\sigma}_{\theta_2}^2$ \textit{w.r.t.} $\mathbf{m}_q$. The sampled $m$ is directly used for computing the attention map.

\textbf{Posterior Networks.}
From the VI view, the posterior network for prototypes is trained to approximate its true posterior distribution given a pair of query samples $(\mathbf{x}_q, \mathbf{y}_q)$ and the support set $\mathcal{S}$. The posterior network for attention maps is trained in a similar way but using only the query pair.
The posterior networks have the same architecture as the prior network of prototypes. However, their outputs $(\boldsymbol{\mu}_{\phi_1},\boldsymbol{\sigma}_{\phi_1}^2) $ come from the aggregation of $k+1$ outputs for $q_{\phi_1}(\mathbf{z}|\mathbf{x}_q,\mathbf{y}_q, \mathcal{S})$ since it has $k+1$ pairs of inputs. Here, the posterior network generates an attention vector $\bm{r}$ about all the $k+1$ input pairs. We then compute the cosine distance $d$ between the attention vector $\bm{r}$ and feature embedding $\bm{e}$ at the pixel level to construct an attention map $\mathbf{m}_q$:
\begin{equation}
\mathbf{m}_{q_{i,j}} = \text{Sigmoid}[d(\bm{e}_{i,j}, \bm{r})].
\end{equation}


\textbf{Segmentation Network.}
Finally, the segmentation network takes the query image $\mathbf{x}_q$, and the sampled prototype vector $\mathbf{z}$, and estimates the attention maps $\mathbf{m}_q$ as inputs to predict the segmentation map $\tilde{\mathbf{y}}_q$, which is the Monte Carlo estimation of the conditional generative distribution $p_{\psi}(\tilde{\mathbf{y}}_q|\mathbf{x}_q,\mathbf{z}, \mathbf{m}_q)$. 
Once we sample an attention map from this distribution, we multiply it by the deep feature of the query image $\mathbf{x}_q$ to enhance a structured embedding. The segmentation net concatenates the attentive embedding and the prototype vector $\mathbf{z}$ sampled from the prior (see  Figure~\ref{fig:fss_vis}) together and produces the output segmentation map:
\begin{equation}
\tilde{\mathbf{y}}_q\sim p_{\psi}(\tilde{\mathbf{y}}_q|\mathbf{x}_q,\mathbf{z}, \mathbf{m}_q).
\end{equation}
At testing time, API generates multiple samples for the prototype and the attention map. This achieves a more accurate prediction using the ensemble of all outputs:
\begin{equation}\label{eq:aggre}
\tilde{\mathbf{y}}_q   = \frac{1}{LM} \sum\limits_{l=1}^{L}  \sum\limits_{j=1}^{M}  f_{\psi}(\mathbf{x}_q \odot \mathbf{m}_q^{(j)}, \mathbf{z}^{(l)}).
\end{equation}

The segmentation network adopts a multi-layer skip-connections structure \cite{ronneberger2015u_unet} to incorporate more spatial information. Besides, there is a CNN-based encoder for the feature embedding shared by the prior, posterior, and segmentation networks. API is an elegant end-to-end framework. All networks are jointly optimized by minimizing the objective (\ref{equa:loss}). The learning algorithm can be summarized in Algorithm \ref{alg:1}. All gradients computations can be efficiently implemented by automatic differentiation tools.

\begin{algorithm}
	\caption{Attentional Prototype Inference}
	\label{alg:1}
	\textbf{Learning:} initialized networks with $\Theta = \{\theta_1, \theta_2, \phi_1, \phi_2, \psi\}$
	
	\KwIn{a training set $D_{\text{train}} = \{S^i,(\mathbf{x}_q^i,\mathbf{y}_q^i)\}_{i=1}^{N_{\text{train}}}$ contains $N_{\text{train}}$ episodes}
	
	\For{\rm{an episode} $\{S,(\mathbf{x}_q,\mathbf{y}_q)\} \in D_{\text{train}}$}
	{
		compute the priors and posteriors by four networks \hfill $\triangleright$ Eq. (6) 
		
		sample prototypes $\mathbf{z}$ from the posterior $q_{\phi_1}(\mathbf{z}|\mathbf{x}_q,\mathbf{y}_q, \mathcal{S}) $    \hfill $\triangleright$ Eq. (7)
		
		sample masks $\mathbf{m}_q$ from the posterior $q_{\phi_2}(\mathbf{m}_q|\mathbf{x}_q,\mathbf{y}_q)$  
		
		MC estimation of conditional prediction $p_{\psi}(\tilde{\mathbf{y}}_q|\mathbf{x}_q,\mathbf{z}, \mathbf{m}_q)$  \hfill $\triangleright$ Eq. (14)
		
		compute gradients of all params. $\Theta$ \textit{w.r.t.} the objective \hfill $\triangleright$ Eq. (5)
		
		back-propagation with gradient descent
	}
	
	\KwOut{the prior networks with $\{\theta_1, \theta_2\}$, SegNet (decoder) with $\psi$}
	\vspace{0.5mm}
	\hrule
	
	\vspace{1mm}
	
	\textbf{Inference}
	
	\KwIn{a query image $\mathbf{x}_q$ and a support set $S$}
	
	compute the priors by two prior networks $\theta_1$ and $\theta_2$
	
	sample the prototype $\mathbf{z}$ from the prior $p_{\theta_1}(\mathbf{z}|S)$  \hfill $\triangleright$ Eq. (7)\\
	
	sample the mask $\mathbf{m}_q$ from the prior $p_{\theta_2}(\mathbf{m}_q|\mathbf{x}_q$  \\
	
	MC estimation of conditional prediction $ p_{\psi}(\tilde{\mathbf{y}}_q|\mathbf{x}_q,\mathbf{z}, \mathbf{m}_q)$  \hfill $\triangleright$ Eq. (14)\\
	\KwOut{segmentation map $\tilde{\mathbf{y}}_q$ for the query image}
	
\end{algorithm}

\section{Experiments}\label{sec:experiments}

\subsection{Datasets and Implementation Details}
We conduct experiments on three commonly used few-shot segmentation benchmarks including PASCAL-${5}^{i}$ \cite{shaban2017one_bmvc}, COCO-${20}^{i}$ \cite{hu2019attention_a-mcg} and FSS-1000 \cite{wei2019fss_fss1000} and one medical imaging dataset of LIDC-IDRI \cite{armato2011lung}. We provide detailed descriptions of these datasets associated with experimental settings as follows. The code for replicating our experiments is available on GitHub (\url{https://github.com/haolsun/API})

\subsubsection{Datasets}
\textbf{PASCAL-${5}^{i}$} originates from PASCAL VOC12 
 and extends annotations. We follow the settings in \cite{yang2020prototype}, splitting the 20 original classes into four folds and conducting cross-validation among them. Specifically, we select 15 classes for training, while the remaining 5 classes are for testing. For a fair comparison, we adopt the same strategy as \cite{yang2020prototype}, randomly sampling 1,000 episodes of support-query pairs for evaluation.

\textbf{COCO-${20}^{i}$} is a challenging dataset built upon MS-COCO 
with 80 object categories. We also divide the 80 classes in MS-COCO into four folds and conduct four-fold cross-validation. Under the same settings as PASCAL-${5}^{i}$, 60 object categories are selected for training, while the remaining 20 categories are used for testing. In each fold, we sample 1000 support-query pairs from the 20 testing classes for evaluation, following \cite{yang2020prototype}. 

\textbf{FSS-1000} is a specialized few-shot segmentation dataset. It contains 1,000 object categories including 520 classes for training, 240 classes for validation, and 240 classes for testing. Following~\cite{wei2019fss_fss1000}, we choose the same 240 categories for testing and train the model on the specified 520 classes with the support of the validation set. The number of testing episodes is 1,000.

\textbf{LIDC-IDRI} is a dataset to represent typical ambiguities in vision signals. It contains 1,018 thoracic CT cases from 1,010 lung patients. The task is to segment lesions of lung nodules given a CT slice. Each CT case has been annotated by 4 individual radiologists that provide segmentation masks for independently detected lesions. We obtain 2,630 sub-cases by extracting a series of slices from 3D volumes and cropping 2D slices to 128 $\times$ 128 pixels centered at the lesion positions. Each sub-case contains a set of 2D images with different cardinalities. Each image corresponds to four annotated masks from independent radiologists (See Figure \ref{fig:example}). For the setting of few-shot segmentation, we consider a sub-case as one episode, \ie, randomly sampling support and query images from one sub-case without replacement. For a total of 2,630 sub-cases, we choose the first 2,030 sub-cases for training and testing the model on the last 300 sub-cases. The remaining 300 sub-cases are for validation. The total numbers of images for training, testing, and validation are 11,631, 1,943, and 1,974.

\begin{figure}[]
	\centering
	\includegraphics[width=0.8\linewidth]{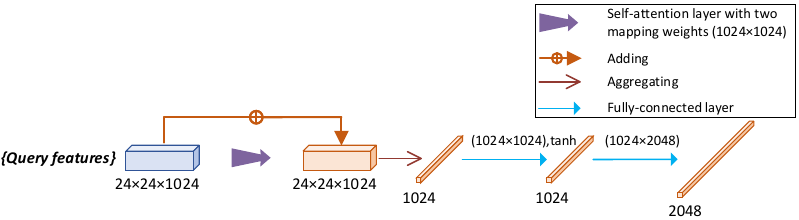}
	\caption{The architecture of prior networks for the attention map.}
	\label{fig:transformer}
\end{figure}

\begin{figure}[]
	\centering
	\includegraphics[width=0.69\linewidth]{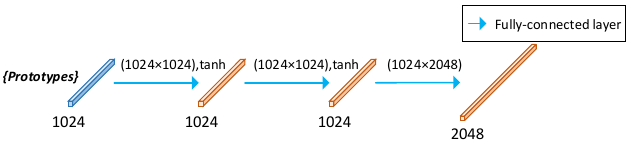}
	\caption{The three networks (\ie, posterior networks for prototypes and masks, prior networks for prototypes) share the same architecture but have different inputs.}
	\label{fig:mlp}
\end{figure}

\begin{figure}[]
	\centering
	\includegraphics[width=0.99\linewidth]{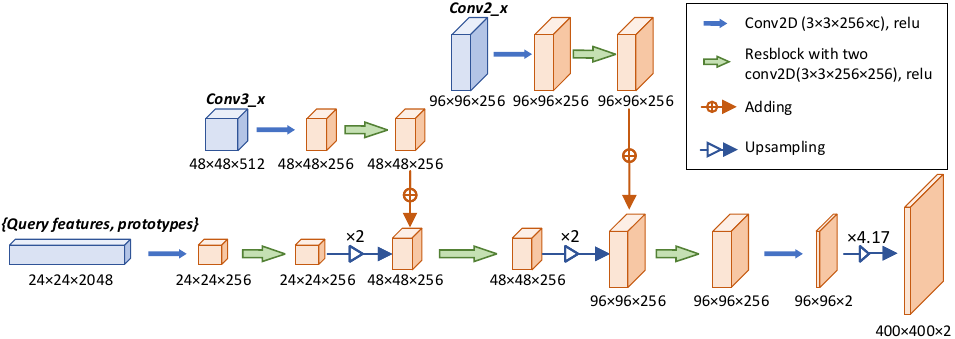}
	\caption{The architecture of the decoder. The light orange cuboid is features from Conv layers. The skipping input is from the corresponding intermediate layers of the pre-trained ResNet.}
	\label{fig:decoder}
\end{figure}

\subsubsection{Implementation Details}
We adopt a ResNet101 
backbone pre-trained on ImageNet 
as the encoder. The decoder is designed as a skip-connection structure \cite{ronneberger2015u_unet}, which is composed of three convolutional blocks to generate segmentation maps. Each block receives the input of the concatenation with the corresponding encoded feature through the skip connections and the decoding features.
The structure of the decoder, prior, and posterior networks are listed in Figures \ref{fig:transformer}, \ref{fig:mlp}, and \ref{fig:decoder}. 
We choose Adam 
as the optimizer and train the model on four NVIDIA Tesla V100 with around 30 epochs. The learning rate is fixed to $5e-7$ for the backbone and $5e-5$ for other layers, and the batch normalization (BN) layers are frozen during training. 
The numbers of samples $L$ and $M$ are set to 10 during the test phase, which is analyzed in detail by our ablation study in Section~\ref{ab}.

We adopt the same metrics as \cite{shaban2017one_bmvc} for evaluation, i.e. Class-IoU (C-IoU) and Binary-IoU (B-IoU). Class-IoU measures the intersection-over-union 
\begin{equation}
\text{IoU} {=} \frac{{\text{TP}}_c}{{\text{TP}}_c + {\text{FP}}_c + {\text{FN}}_c}
\end{equation}
 for each class, where TP, FP and FN are the number of pixels that are true positives, false positives and false negatives of the predicted segmentation masks for each foreground category $c$. Binary-IoU measures the IoU between the foreground and background pixels, where all object classes are treated as foreground.

We generalize the IoU metric to the cross energy distance (CED) on the LIDC-IDRI dataset to measure the distance between two distributions. The core idea of CED is leveraging distances between observations. Let $\mathbf{y}, \hat{\mathbf{y}}$ denote the annotation and the prediction, respectively. $d(\cdot)$ is the distance between two observations. For $m$ independent annotation samples and $n$ prediction samples, we have
\begin{equation}
	D_{\text{CED}} = \text{E}[d(\mathbf{y}, \hat{\mathbf{y}}) ] = \frac{1}{mn} \sum_{i=1}^{m} \sum_{j=1}^{n} d(\mathbf{y}_i, \hat{\mathbf{y}}_j).
	\label{equa:ced}
\end{equation}
Here, we choose $d(\mathbf{y}, \hat{\mathbf{y}}) = 1- \text{IoU}(\mathbf{y}, \hat{\mathbf{y}})$. A smaller CED value indicates that the prediction distribution is close to the ground truth distribution and can characterize the ambiguity that appears in images. In our LIDC-IDRI experiments, we set $m=4$ since there are four annotation masks for one image.

\subsection{Comparison with State-of-the-Arts}

\begin{table}[]
	\centering
	\caption{Comparison in terms of Class-IoU and Binary-IoU on PASCAL-$5^i$.}
	\label{tab:pascal}
	\begin{tabular}{ll cc cc}
		\toprule
		\centering
		&&\multicolumn{2}{c}{\textbf{C-IoU}}&\multicolumn{2}{c}{\textbf{B-IoU}}\\
		\cmidrule(lr){3-4} \cmidrule(lr){5-6}
		Method&Backbone&\textit{1-shot}&\textit{5-shot}&\textit{1-shot}&\textit{5-shot}\\
		\midrule
		PMM~\cite{yang2020prototype}&ResNet50&56.3& - &57.3&-\\
		CRNet~\cite{liu2020crnet}&ResNet50& 55.7& -&58.8&-\\
		FS-PARN~\cite{LI2021107882}&ResNet50&53.7& 67.9 &58.3&72.6\\
		FWB~\cite{nguyen2019feature_boost}&ResNet101& 56.2 &&59.9&-\\
		\textbf{API~}(\textit{Ours}) & ResNet101 & \textbf{57.4} &  \textbf{71.4}& \textbf{60.7}& \textbf{73.2}\\
		
		\hline
		HSNet~\cite{Min_2021_ICCV}& ResNet101 &66.2 &72.5 & 70.4 & 80.6\\
		VAT~\cite{hong2022cost} & ResNet101 &\textbf{67.5} &\textbf{78.8} &71.6 & \textbf{82.0 }\\
		DGPNet~\cite{johnander2022dense}& ResNet101 & 64.8& - &\textbf{75.4} &-\\
		SSP~\cite{fan2022self}&ResNet101 &64.6&-&73.1&-\\
		\textbf{API*~}& ResNet101& 64.7&76.3&70.2&80.4 \\
		
		\bottomrule
	\end{tabular}
\end{table}

\subsubsection{Performance on PASCAL-${5}^{i}$}
In Table~\ref{tab:pascal}, we compare the performance of API with typical methods on PASCAL-${5}^{i}$ in terms of the Class-IoU metric and Binary-IoU. API outperforms those prototype-based methods by good margins under both the 1-shot and 5-shot settings (57.4\%, 60.7\%). The 1-shot setting is more challenging than the 5-shot setting due to the much larger intra-class variation. The Monte Carlo estimation in our probabilistic model serves as an ensemble of the prediction results. Specifically, API outperforms PMM which computes segmentation masks with multiple prototypes. This accounts for the robustness of our API in the 1-shot case. We also evaluate the model in terms of Binary-IoU. Our model again yields comparable performance under both the 1-shot and 5-shot settings of $70.3\%$ and $72.1\%$.

However, due to the intrinsic frailty of prototype-based methods, API has limited capability of handling objects with small sizes or complex shapes. Recall the computation of the prototype is from feature maps at the high level, discriminative information of objects with small sizes may vanish as downsampling operations, especially for small objects with complex backgrounds. This challenge also exists when a large object has complex shapes, \ie some parts of the object, such as table legs, might be omitted. Hierarchical prototypes would be potential to tackle it as we discussed in Conclusion. Therefore, we remove one class with the lowest confidence in each fold and demote it as API*. As shown in the bottom part of Table~\ref{tab:pascal}, API* can achieve considerable performance compared with the state-of-the-art, \eg, $64.7\%$ for the setting of 1-shot. Compared with other methods, our prototype-based model is more efficient and easy to be implemented. Because of the advanced architecture of transformers, VAT~\cite{hong2022cost} obtained the highest IoU metric for most cases on PASCAL-${5}^{i}$.

\begin{figure*}[t]
	\centering
	\includegraphics[width=\linewidth]{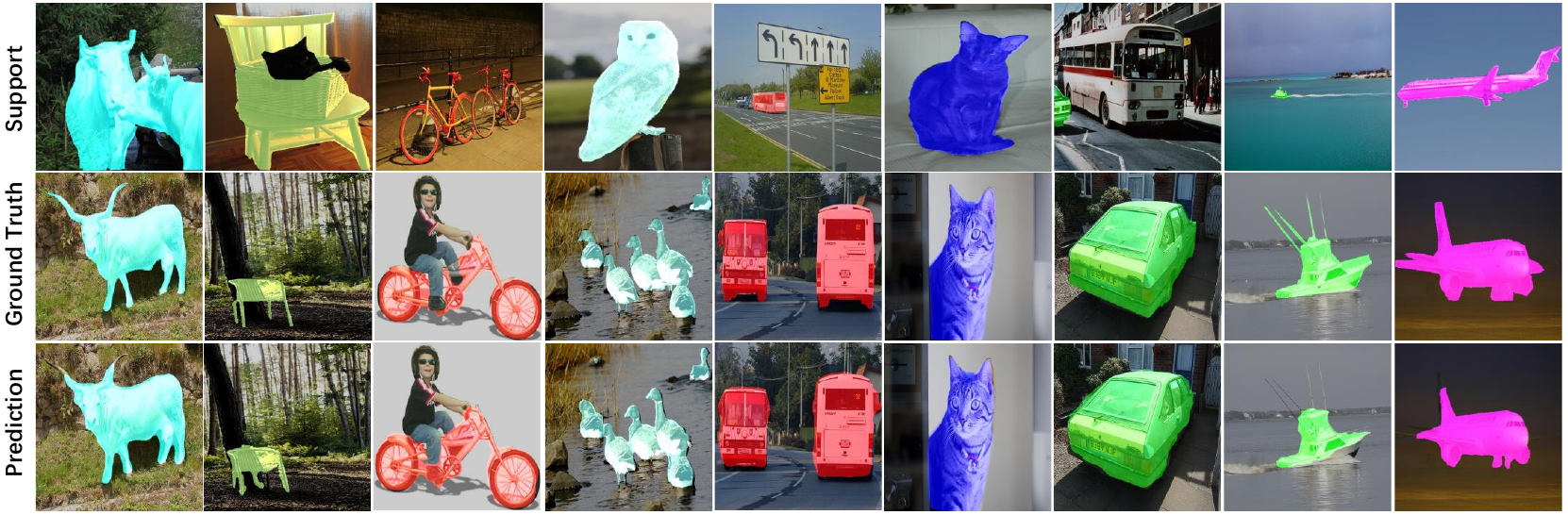}
	\caption{Visualization of 1-shot segmentation results on  PASCAL-${5}^{i}$. From top to bottom are support image, ground-truth and prediction. Our method achieves accurate segmentation maps for query images, even in cases of considerable variation of objects from support to query images.}
	\label{fig:voc_vis}
\end{figure*}

Some qualitative results on PASCAL-${5}^{i}$ are visualized in Figure~\ref{fig:voc_vis}. The proposed API is capable of producing more accurate segmentation under various challenging scenarios, where the query images vary in appearance and object size from the associated support images. For instance, in the fifth column, the size and viewpoint of the bus in the query image 
is significantly different from the annotated plane in the support image; in the eighth column, the annotated boat in the support image is much smaller than the one in the query image.

\begin{table}[]
	\centering
	\caption{Comparison in terms of Class-IoU and Binary-IoU on COCO-$20^i$.}
	\label{tab:coco}
	\begin{tabular}{ll cc cc}
		\toprule
		\centering
		&&\multicolumn{2}{c}{\textbf{C-IoU}}&\multicolumn{2}{c}{\textbf{B-IoU}}\\
		\cmidrule(lr){3-4} \cmidrule(lr){5-6}
		Method&Backbone&\textit{1-shot}&\textit{5-shot}&\textit{1-shot}&\textit{5-shot}\\
		\midrule
		PANet~\cite{wang2019panet_panet}&VGG16&20.9& 29.7 &59.2&\textbf{63.0}\\
		PMM~\cite{yang2020prototype}&ResNet50& 30.6& 35.5 &-&-\\
		FS-PARN~\cite{LI2021107882}&ResNet50& 29.5& 36.2 &-&-\\
		A-MCG~\cite{hu2019attention_a-mcg}&ResNet101&-&-&52.0&54.7\\
		FWB~\cite{nguyen2019feature_boost}&ResNet101& 21.2 &23.7&-&-\\
		\textbf{API~}(\textit{Ours}) & ResNet101 & \textbf{36.3} &  \textbf{41.0}& \textbf{61.9}& 62.7\\
		
		\hline
		HSNet~\cite{Min_2021_ICCV}& ResNet101 &41.2 &49.5 & \textbf{69.1} & \textbf{72.4}\\
		VAT~\cite{hong2022cost} & ResNet101 &41.3 &47.9 &68.8 & \textbf{72.4 }\\
		DGPNet~\cite{johnander2022dense}& ResNet101 & \textbf{46.7}& \textbf{57.9} &- &-\\
		SSP~\cite{fan2022self}&ResNet101 &42.0&50.2&-&-\\
		\textbf{API*~}& ResNet101& 41.8&49.4&68.2&70.8 \\
		
		\bottomrule
	\end{tabular}
\end{table}

\subsubsection{Performance on COCO-${20}^{i}$}

\begin{figure*}[]
	\centering
	\includegraphics[width=\linewidth]{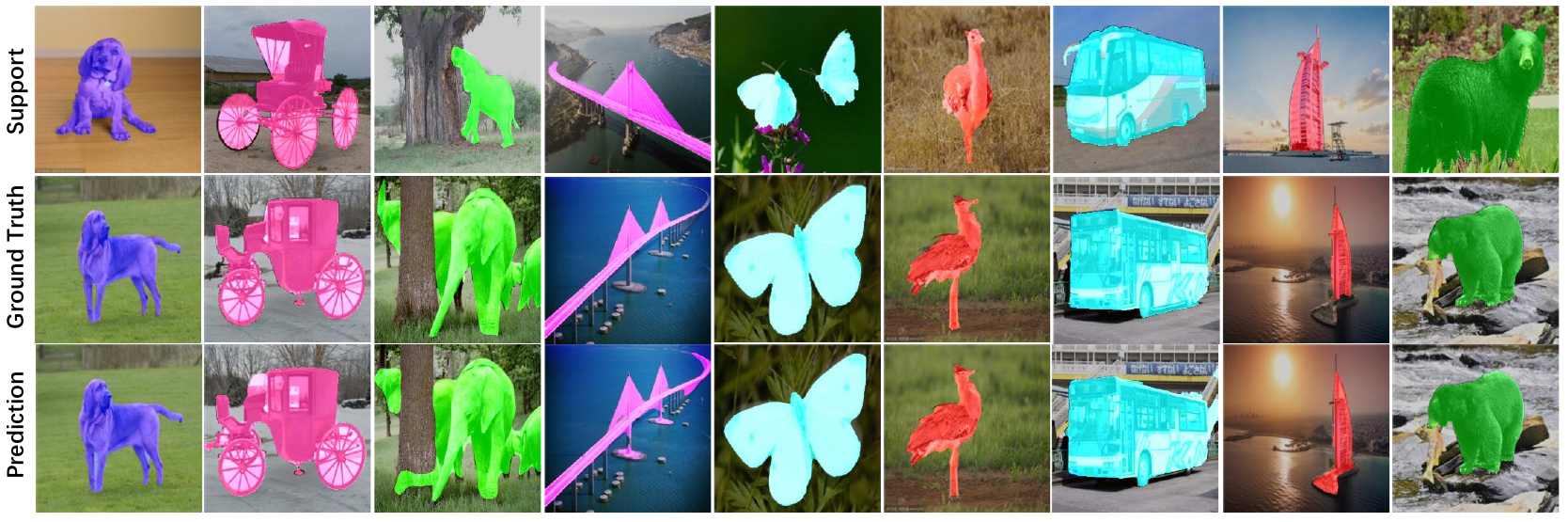}
	\caption{Visualization of 1-shot segmentation results on  COCO-${20}^{i}$. Our method successfully predicts the segmentation maps for query images, though the objects in the query image are considerably different from those in the support images in terms of appearance, size and viewpoints.}
	\label{fig:coco_vis}
\end{figure*}

COCO-${20}^{i}$ is more challenging than PASCAL-${5}^{i}$ since the scenes in COCO-${20}^{i}$ are more complex with more intra-class diversity. Therefore, few-shot segmentation on COCO-${20}^{i}$ has more ambiguity and it is difficult to acquire an effective class-specific prototype. As can be seen in Table~\ref{tab:coco}, our method outperforms the homogeneous method PMM~\cite{yang2020prototype} by $5.7\%$ and $5.5\%$ in terms of Class-IoU under the 1-shot and 5-shot settings. Since the COCO-${20}^{i}$ dataset contains more complex objects than PASCAL-${5}^{i}$, we drop five classes with the lowest confidence in each fold. As shown in Table~\ref{tab:coco}, we obtain reasonable results on the rest of 15 classes in each fold, e. g., $41.8\%$ of Class-IoU. As DGPNet~\cite{johnander2022dense} applied the powerful probabilistic model of Gaussian Process and introduced extra mask encoders, it non-trivially outperforms those recent works. The qualitative results are provided in Figure~\ref{fig:coco_vis}. Our method successfully predicts the segmentation maps for query images, though the objects in the query image are significantly different from those in the support images in  terms of appearance, size, and viewpoints. As shown in the third column, the object can also be successfully segmented even with serve occlusions.

\begin{table}[]
	\centering
	\caption{Comparison in terms of Positive-IoU on FSS-1000.}
	\label{tab:fss}
	\begin{tabular}{llcc}
		\toprule
		Method&Backbone &\textit{1-shot}&\textit{5-shot}\\
		\midrule
		OSLSM~\cite{shaban2017one_bmvc}&VGG16 &70.3&73.0\\
		FSS-1000~\cite{wei2019fss_fss1000}&VGG16 &73.5& 80.1 \\
		\textbf{API} (\textit{This paper}) & VGG16 & 83.4&  85.3\\
		\textbf{API} (\textit{This paper}) & ResNet101 & \textbf{85.6}&  \textbf{88.0}\\
		\bottomrule
	\end{tabular}
\end{table}

\subsubsection{Performance on FSS-1000}
\label{subsubsec:fss}
\begin{figure*}[]
	\centering
	\includegraphics[width=\linewidth]{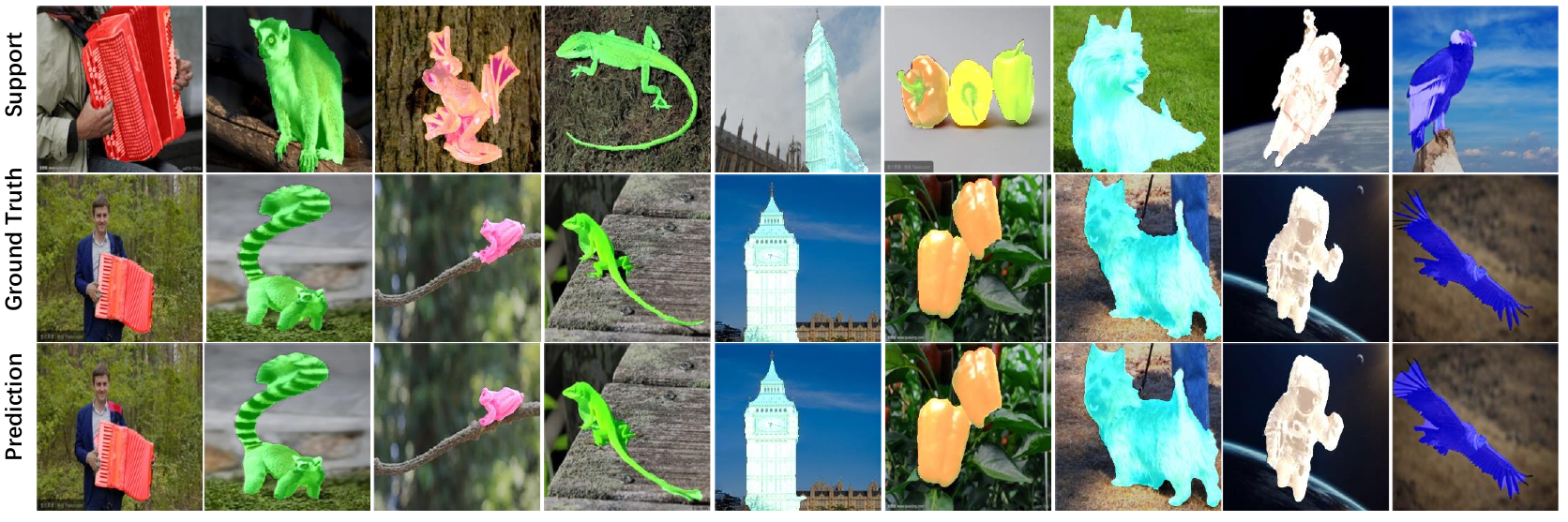}
	\caption{Visualization of 1-shot segmentation results on FSS-1000. Our method is able to accurately segment objects in challenging scenarios, including those with variants in background and appearance.}
	\label{fig:fss_vis}
	
\end{figure*}

We evaluate our method following the official
evaluation protocols in~\cite{wei2019fss_fss1000}. The evaluation metric used for FSS-1000 is the IoU of positive labels in a binary segmentation map. Since this dataset contains the validation set, we select the prediction model with the support of the validation set (i. e., the model for the best validation class-IoU). Then, we report the result on the testing set. A performance comparison with other models in terms of Positive-IoU (P-IoU) is provided in Table~\ref{tab:fss}. Our method improves over the state-of-the-art set by Wei et al.~\cite{wei2019fss_fss1000} by $12.1\%$ and $7.9\%$ in the 1-shot and 5-shot settings, demonstrating the effectiveness of our proposal for few-shot segmentation 
across large-scale semantic categories. Figure~\ref{fig:fss_vis} visualizes segmentation results on the FSS-1000 dataset, where API produces accurate segmentation maps close to the ground truth. As shown in Figure~\ref{fig:fss_vis}, the foreground object is usually located at the center of the image and distinct from the background. This could lead to considerable performance for our model on FSS-1000.

\subsubsection{Performance on LIDC-IDRI}

To study the capability of handling ambiguities in few-shot segmentation, we conduct experiments on LIDC-IDRI with typical image ambiguities. The evaluation metric for LIDC-IDRI is the cross energy distance (CED) rather than IoU values. We compared our API with two state-of-the-art deterministic models, \ie, FS-PARN~\cite{LI2021107882} and HSNet~\cite{Min_2021_ICCV}. Since it is unfeasible to directly obtain prediction samples, we adopt an ensemble strategy that trains a set of predictors. Considering there are 4 annotations, we initialize 15 models and select different combinations of annotations to train the model. Thus, the total number of those combinations is $C_4^1 + C_4^2 + C_4^3 + C_4^4 = 15$. For each model, we train it with the data batch that randomly sampled annotation masks from the current combination. Once 15 predictors are obtained, we randomly select a subset of predictors to compute multiple segmentation maps for testing. As for our API model, we randomly choose one of 4 annotation masks at each update step. In this experiment, the sampling number $n$ in CED (Eq. \ref{equa:ced}) is set as 9 for all methods. We train the model on the training set of 2,030 sub-cases with the support of the validation set of 300 sub-cases. The results shown in Table \ref{tab:lidc} are CED values on the testing set. We test the model five times with different predictors selected randomly and report the average value for evaluation. Our API achieves non-trivial improvement in terms of CED compared with state-of-the-art methods. In contrast to evaluating the method with the class-IoU metric, the results of CED include all prediction samples and annotation masks, demonstrating the good statistical property of our proposed framework.

\begin{table}[]
	\centering
	\caption{Comparison in terms of CED ($\%$) for 1-shot on LIDC-IDRI. Smaller is better.}
	\label{tab:lidc}
	\begin{tabular}{l|ccc}
		\toprule
		Backbone& API &FS-PARN~\cite{LI2021107882} & HSNet~\cite{Min_2021_ICCV}\\
		\midrule
		ResNet101 &85$\pm$0.1 &94$\pm$1.1 & 91$\pm$1.2\\
		
		\bottomrule
	\end{tabular}
\end{table}

\subsection{Ablation Study}


\subsubsection{Benefit of Probabilistic Modeling}
\label{ab}
Different from previous deterministic models by learning a deterministic prototype vector, our probabilistic methods infers the distributions of the class prototype and the attention vector for each image. To demonstrate the advantage of the proposed probabilistic modeling, we implement a deterministic counterpart. We utilize the same network architecture for fair comparison and predict a deterministic class prototype vector or an attention map by the $\boldsymbol{\mu}$ branch and remove the KL divergence term during training. We implement both models with a VGG-16 
, ResNet50, and ResNet101 
backbone, which are commonly adopted in previous works \cite{nguyen2019feature_boost,zhang2019pyramid}.

The results on PASCAL-$5^i$ are shown in Table~\ref{tab:baseline_re}. Our attentional prototype inference consistently achieves better performance than the deterministic models under both the $1$ and $5$-shot settings in terms of Class-IoU and the Binary-IoU metrics, because the proposed probabilistic modeling of prototypes and attention maps is more expressive of object classes and has a compelling capability of capturing the categorical concepts of objects. Therefore, the learned model is endowed with a stronger generalization ability for query images that usually exhibits large variations.
The results also illustrate the advantage of probabilistic modeling for few-shot segmentation. As the ResNet101 backbone outperforms both VGG16 and ResNet50, we adopt ResNet101 as the backbone network in our experiments.

\begin{table*}
	\setlength{\tabcolsep}{6pt}
	\centering
	\caption{
		Benefit of probabilistic modeling on PASCAL-$5^i$. The proposed probabilistic modeling shows consistent advantages over deterministic models in terms of Class-IoU and Binary-IoU with different backbone networks and under both the 1-shot and 5-shot settings.}
	\setlength{\tabcolsep}{0.7mm}
	\begin{tabular}{l cccc cccc cccc}
		\toprule
		& \multicolumn{4}{c}{\textbf{VGG}} &
		\multicolumn{4}{c}{\textbf{ResNet50}} &
		\multicolumn{4}{c}{\textbf{ResNet101}}\\ 
		\cmidrule(lr){2-5} \cmidrule(lr){6-9} \cmidrule(lr){10-13}
		& \multicolumn{2}{c}{\textbf{C-IoU}} & \multicolumn{2}{c}{\textbf{B-IoU}} &
		\multicolumn{2}{c}{\textbf{C-IoU}} &
		\multicolumn{2}{c}{\textbf{B-IoU}} &
		\multicolumn{2}{c}{\textbf{C-IoU}}&\multicolumn{2}{c}{\textbf{B-IoU}}\\
		\cmidrule(lr){2-3} \cmidrule(lr){4-5} \cmidrule(lr){6-7}\cmidrule(lr){8-9} \cmidrule(lr){10-11}\cmidrule(lr){12-13}
		\textit{k-shot}& 1 & 5 & 1 & 5 & 1 & 5 & 1 & 5 & 1 & 5 & 1 & 5\\
		\midrule
		Deter. & 51.6 & 53.1 & 64.1 & 65.3 & 54.1 & 57.4 & 65.7 & 68.9 & 55.4 & 58.7 & 66.3 & 69.9\\
		\textbf{API}  & 52.7 & 55.6 & 65.2 & 66.0 & 55.9 & 59.8 & 69.3 & 71.7 & 57.4 & 60.7 & 71.3 & 73.2\\
		\bottomrule
	\end{tabular}
	\label{tab:baseline_re}
\end{table*}

\subsubsection{Effectiveness of Latent Attention Mechanism}
The newly introduced attention mechanism faithfully leverages the structure information that is neglected by the prototype, which is essential for mask prediction. The transformer architecture adopted in our model can utilize the context knowledge from pixels with high fidelity and enhances the foreground to achieve accurate prediction. We conduct extensive comparison experiments with the baseline variational prototype inference (VPI)~\cite{wang2021variational} to show the effectiveness of the latent attention mechanism. As shown in Table \ref{tab:vpi_comp}, we evaluate their performance on three benchmarks of PASCAL-${5}^{i}$, COCO-${20}^{i}$ and FSS-1000 with three metrics of Class-IoU, Binary-IoU, and Positive IoU. API achieves considerable improvement compared with VPI in most cases. In particular, API improves VPI by $7.8\%$ under the 1-shot, and $8.1\%$ under the 5-shot settings for Class-IoU on COCO-20$^i$. The performance advantage of API in Binary-IoU compared to VPI is relatively smaller than in Class-IoU on COCO-20$^i$. This is due to the bias of Binary-IoU towards objects that cover a large part of the foreground and background areas.

\begin{table*}[]
	\centering
	\makeatletter\def\@captype{table}\makeatother\caption{Comparison of the homogeneous method on three benchmarks. }
	\setlength{\tabcolsep}{0.5mm}
	\begin{tabular}{l  cc  cc c}
		\toprule
		&\multicolumn{4}{c}{PASCAL-5$^i$   }&FSS-1000\\ 
		&\multicolumn{2}{c}{\textbf{C-IoU}}&\multicolumn{2}{c}{\textbf{B-IoU}}&\textbf{P-IoU}\\%
		\cmidrule(lr){2-3} \cmidrule(lr){4-5} \cmidrule(lr){6-6}
		&\textit{1-shot}&\textit{5-shot}&\textit{1-shot}&\textit{5-shot}&\textit{1-shot}\\\hline
		VPI&{57.3}&{60.4}&70.3&72.1&84.3\\
		\textbf{API} &\textbf{57.4}&\textbf{60.7}&\textbf{71.3} &\textbf{73.2}&\textbf{85.6}\\
		\midrule
		&\multicolumn{4}{c}{COCO-20$^i$   }&FSS-1000\\ 
		&\multicolumn{2}{c}{\textbf{C-IoU}}&\multicolumn{2}{c}{\textbf{B-IoU}}&\textbf{P-IoU}\\%
		\cmidrule(lr){2-3} \cmidrule(lr){4-5} \cmidrule(lr){6-6}
		&\textit{1-shot}&\textit{5-shot}&\textit{1-shot}&\textit{5-shot}&\textit{1-shot}\\\hline
		VPI&{23.4}&27.8&61.1&\textbf{63.0}&87.7\\
		\textbf{API} &\textbf{31.2}&\textbf{35.9}&\textbf{61.4}&62.3&\textbf{88.0}\\
		
		\bottomrule
	\end{tabular}
	\label{tab:vpi_comp}
\end{table*}

\begin{figure*}[]
	\centering
	\includegraphics[width=\linewidth]{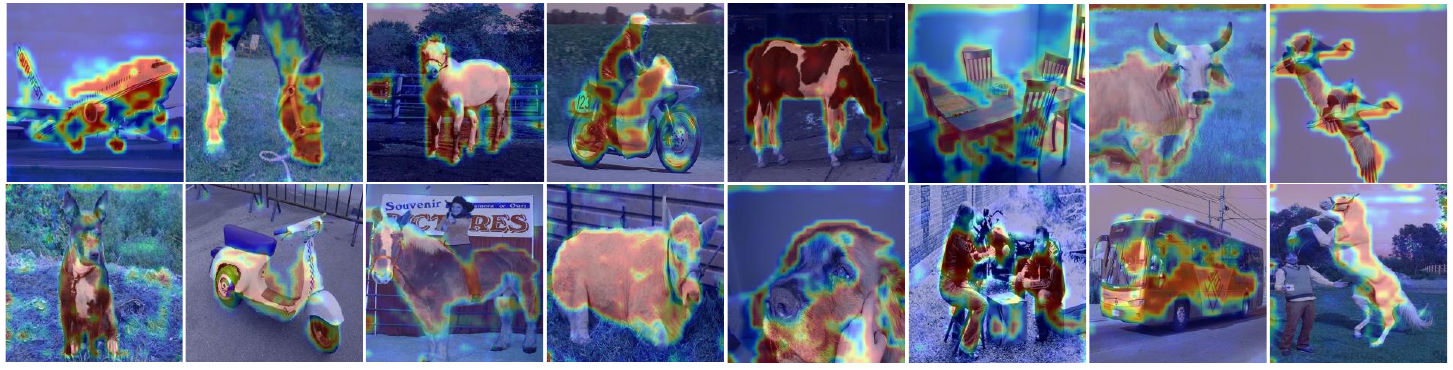}
	\caption{Visualization of the attention maps. The foreground object regions are highlighted, which helps achieve better prediction of segmentation maps.}
	\label{fig:att_vis}
	\vspace{-3mm}
\end{figure*}

We also visualize the attention maps computed by our API model in Figure~\ref{fig:att_vis}. The pixel on the foreground object is highlighted to enhance the prediction of the prediction maps. The proposed latent attention module can capture the outline of the foreground objects. Besides, this module merely increases a little inference time. As evidenced in Figure \ref{fig:speed}, when we fix $L=10$ and increase $M$ from 1 to 15, the extra time introduced by the attention module is less than 0.1 seconds even under the 5-shot setting. These observations indicate the advantage of our strategy for computational efficiency. 

\subsubsection{Effect of Monte Carlo Sampling}
\begin{figure*}[]
	\begin{minipage}{.48\textwidth}
		\centering
		\includegraphics[width=0.85\linewidth]{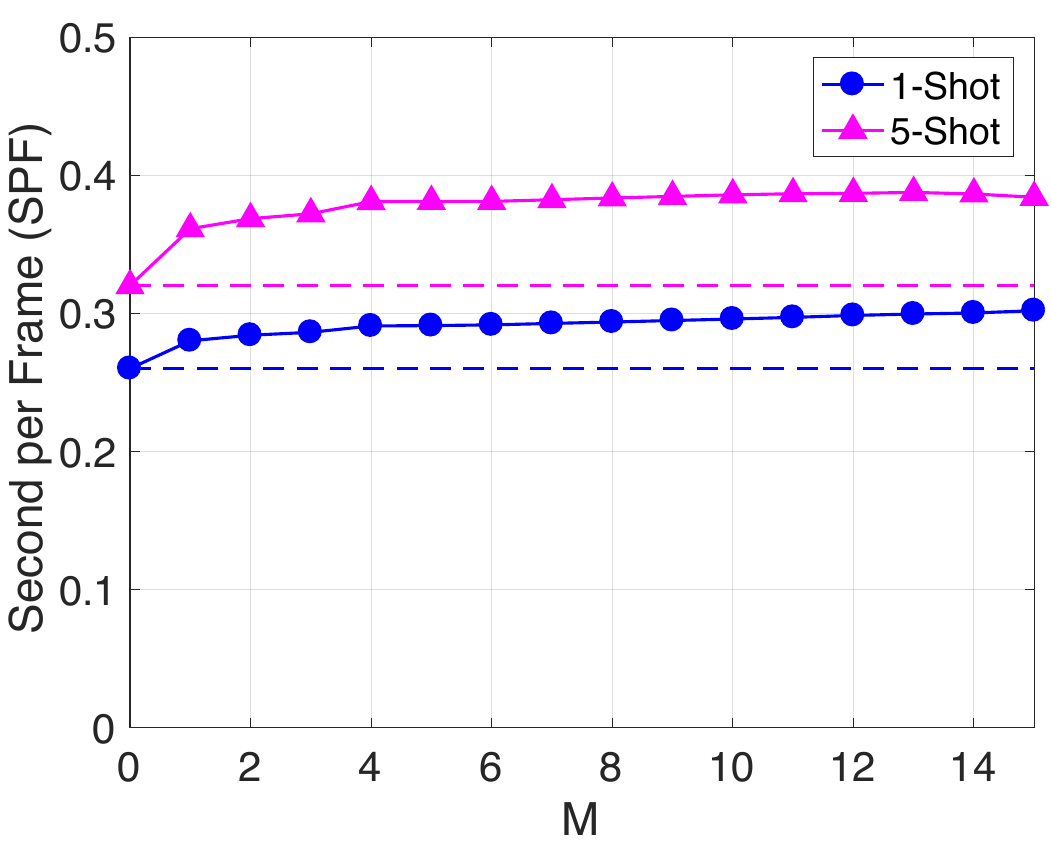}
		\vspace{-3mm}
		\caption{The attention mechanism increases the inference time slightly, less than 0.1 second under the 5-shot setting with $M=15$.}
		\label{fig:speed}
	\end{minipage}
	\hspace{3mm}
	\begin{minipage}{.48\textwidth}
		\centering
		\includegraphics[width=\linewidth]{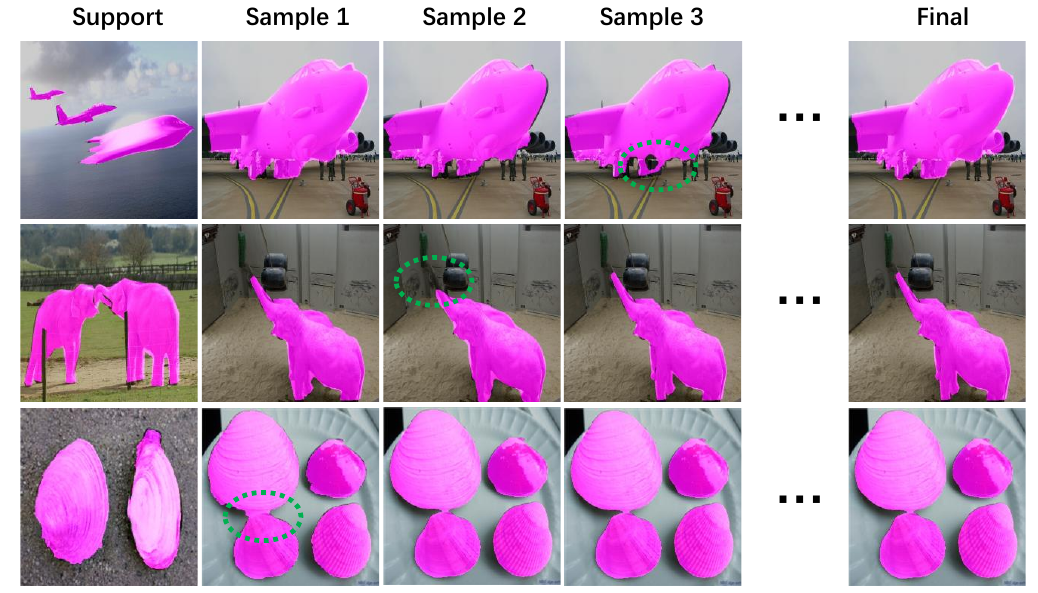}
		\caption{Effect of Monte Carlo Sampling. Segmentation maps produced by sampled individual prototypes tend to be noisy, but the final segmentation maps by the aggregated prototype have less noise. }
		\label{fig:sample_vis}
	\end{minipage}
\end{figure*}

The segmentation map is estimated by Monte Carlo sampling that obtains multiple prototypes $\textbf{z}$ and attention maps $\textbf{m}$, and produces multiple outputs, then, all outputs are aggregated to produce the final segmentation map. We conduct a qualitative analysis of the effect of the Monte Carlo sampling on the segmentation results. As shown in Figure~\ref{fig:sample_vis}, the segmentation map for each sampled prototype is not always adequate. For example, in the first row, the segmentation map generated by 3 times sampling does not completely recover the object. By averaging the segmentation maps produced by the individual samples, the final segmentation map tends to be more complete and robust. 

The quantitative results (Class-IoU of the sample number 5 \& 15: 0.56 vs. 0.57) show that the prediction result turns to be better as the number of samples increases. Despite the fact that the segmentation results are more accurate given more samples, it will take more time for inference. We observe that the performance tends to saturate when $L, M$ reach $15$. Therefore, in our experiments, we set $L, M$ to 5 during inference to achieve precise segmentation maps with acceptable inference cost.

\section{Discussion and Conclusion}\label{sec:conclusions}

This paper tackles few-shot segmentation from a probabilistic perspective. It contains two latent variables from different levels: 1) a global latent variable is inferred from the data that represents the prototype of each object category; 2) a local latent variable is designed to represent an attention map for each image, which highlights the foreground object for improved segmentation. We develop attentive prototype inference (API) to leverage variational inference for efficient optimization. By probabilistic modeling, API is capable of enhancing its generalization ability by handling the inherent uncertainty caused by limited
data and the intra-class variations of objects, which is essential for generalizing to new unseen categories in experimental practice.  

The inherent property of the prototype-based method is computing the foreground prototype with the guide of label masks at the image level. Those prototypes are usually ambiguous for foreground objects at the feature level, especially for instances with small sizes. This could be resolved by taking hierarchical prototypes of varied feature levels into account in future work, \ie, computing prototypes at the different level of the encoder, especially, those that are close to the input layer.

The probabilistic manner is essential in API. The key step is the Monte Carlo (MC) estimation for the segmentation prediction. To achieve a better estimation result, we usually consider sampling multiple hypotheses. This would bring extra computational costs, especially during the training stage. However, it is unnecessary to obtain an accurate MC estimation of the conditional prediction distribution at the training stage. Recall SGD in neural network optimization, the biased estimation could introduce randomness for the learning process to achieve better generalization. Therefore, we choose the sampling number of 1 at the training stage. Also, there is an extra benefit that can be considered as the regularization by injecting uncertainty into the learning process. Besides, the inference speed at the testing stage can be acceptable as the sampling number of the MC estimation is small (i.e., 10 in all experiments). 

The probabilistic formulation could handle ambiguities in the segmentation task by producing a set of diverse but plausible segmentation results. It could potentially have a high impact on clinical applications. Multiple segmentation
hypotheses from our model could provide diagnosis probabilities or guide steps to resolve ambiguities. 

\section*{Acknowledgments}
This research was supported in part by Natural Science Foundation of China (Nos. 62106129, 62176139, 62106128), Natural Science Foundation of Shandong Province (Nos. ZR2021QF053, ZR2021QF001), Major Basic Research Project of Natural Science Foundation of Shandong Province (No. ZR2021ZD15), The Fundamental Research Funds of Shandong University and The Open Research Project Programme of the State Key Laboratory of Internet of Things for Smart City (University of Macau) (No. SKL-IoTSC(UM)-2021-2023/ORP/GA05/2022).

\bibliography{refs}

\begin{thebibliography}{10}
\expandafter\ifx\csname url\endcsname\relax
  \def\url#1{\texttt{#1}}\fi
\expandafter\ifx\csname urlprefix\endcsname\relax\def\urlprefix{URL }\fi
\expandafter\ifx\csname href\endcsname\relax
  \def\href#1#2{#2} \def\path#1{#1}\fi

\bibitem{brostow2009semantic}
G.~J. Brostow, J.~Fauqueur, R.~Cipolla, Semantic object classes in video: A
  high-definition ground truth database, Pattern Recognition Letters 30~(2)
  (2009) 88--97.

\bibitem{chen2017deeplab}
L.-C. Chen, G.~Papandreou, I.~Kokkinos, K.~Murphy, A.~L. Yuille, Deeplab:
  Semantic image segmentation with deep convolutional nets, atrous convolution,
  and fully connected crfs, IEEE Transactions on Pattern Analysis and Machine
  Intelligence 40~(4) (2017) 834--848.

\bibitem{long2015fully}
J.~Long, E.~Shelhamer, T.~Darrell, Fully convolutional networks for semantic
  segmentation, in: Computer Vision and Pattern Recognition, 2015.

\bibitem{fan2022self}
Q.~Fan, W.~Pei, Y.-W. Tai, C.-K. Tang, Self-support few-shot semantic
  segmentation, in: European Conference on Computer Vision, Springer, 2022, pp.
  701--719.

\bibitem{luo2022meta}
S.~Luo, Y.~Li, P.~Gao, Y.~Wang, S.~Serikawa, Meta-seg: A survey of
  meta-learning for image segmentation, Pattern Recognition (2022) 108586.

\bibitem{rosch1973natural}
E.~H. Rosch, Natural categories, Cognitive psychology 4~(3) (1973) 328--350.

\bibitem{snell2017prototypical}
J.~Snell, K.~Swersky, R.~Zemel, Prototypical networks for few-shot learning,
  in: Neural Information Processing Systems, 2017, pp. 4077--4087.

\bibitem{li2021adaptive}
G.~Li, V.~Jampani, L.~Sevilla-Lara, D.~Sun, J.~Kim, J.~Kim, Adaptive prototype
  learning and allocation for few-shot segmentation, in: Computer Vision and
  Pattern Recognition, 2021, pp. 8334--8343.

\bibitem{dong2018few_pl}
N.~Dong, E.~Xing, Few-shot semantic segmentation with prototype learning., in:
  British Machine Vision Conference, 2018.

\bibitem{ZHOU2022108290}
Q.~Zhou, X.~Wu, S.~Zhang, B.~Kang, Z.~Ge, L.~{Jan Latecki}, Contextual ensemble
  network for semantic segmentation, Pattern Recognition 122 (2022) 108290.

\bibitem{lu2021segmenting}
X.~Lu, W.~Wang, J.~Shen, D.~Crandall, L.~Van~Gool, Segmenting objects from
  relational visual data, IEEE Transactions on Pattern Analysis and Machine
  Intelligence~(01) (2021) 1--1.

\bibitem{zhao2017pyramid}
H.~Zhao, J.~Shi, X.~Qi, X.~Wang, J.~Jia, Pyramid scene parsing network, in:
  Computer Vision and Pattern Recognition, 2017, pp. 2881--2890.

\bibitem{zhang2021gpnet}
Y.~Zhang, X.~Sun, J.~Dong, C.~Chen, Q.~Lv, Gpnet: gated pyramid network for
  semantic segmentation, Pattern Recognition 115 (2021) 107940.

\bibitem{shaban2017one_bmvc}
A.~Shaban, S.~Bansal, Z.~Liu, I.~Essa, B.~Boots, One-shot learning for semantic
  segmentation, in: British Machine Vision Conference, 2017.

\bibitem{siam2019amp_amp}
M.~Siam, B.~N. Oreshkin, M.~Jagersand, Amp: Adaptive masked proxies for
  few-shot segmentation, in: International Conference on Computer Vision, 2019,
  pp. 5249--5258.

\bibitem{okazawa2022interclass}
A.~Okazawa, Interclass prototype relation for few-shot segmentation, in:
  European Conference on Computer Vision, Springer, 2022, pp. 362--378.

\bibitem{yang2020prototype}
B.~Yang, C.~Liu, B.~Li, J.~Jiao, Q.~Ye, Prototype mixture models for few-shot
  semantic segmentation, in: European Conference on Computer Vision, 2020, pp.
  763--778.

\bibitem{tian2020pfenet}
Z.~Tian, H.~Zhao, M.~Shu, Z.~Yang, R.~Li, J.~Jia, Prior guided feature
  enrichment network for few-shot segmentation, IEEE Transactions on Pattern
  Analysis and Machine Intelligence~(01) (2020) 1--1.

\bibitem{LI2021107882}
Y.~Li, P.~Zhang, X.~Xu, Y.~Lai, F.~Shen, L.~Chen, P.~Gao, Few-shot prototype
  alignment regularization network for document image layout segementation,
  Pattern Recognition 115 (2021) 107882.

\bibitem{singh2021metamed}
R.~Singh, V.~Bharti, V.~Purohit, A.~Kumar, A.~K. Singh, S.~K. Singh, Metamed:
  Few-shot medical image classification using gradient-based meta-learning,
  Pattern Recognition 120 (2021) 108111.

\bibitem{Min_2021_ICCV}
J.~Min, D.~Kang, M.~Cho, Hypercorrelation squeeze for few-shot segmentation,
  in: International Conference on Computer Vision, 2021, pp. 6941--6952.

\bibitem{hong2022cost}
S.~Hong, S.~Cho, J.~Nam, S.~Lin, S.~Kim, Cost aggregation with 4d convolutional
  swin transformer for few-shot segmentation, in: European Conference on
  Computer Vision, Springer, 2022, pp. 108--126.

\bibitem{johnander2022dense}
J.~Johnander, J.~Edstedt, M.~Felsberg, F.~S. Khan, M.~Danelljan, Dense gaussian
  processes for few-shot segmentation, in: European Conference on Computer
  Vision, Springer, 2022, pp. 217--234.

\bibitem{kingma2013auto}
D.~P. Kingma, M.~Welling, Auto-encoding variational bayes, in: International
  Conference on Learning Representations, 2014.

\bibitem{lim2021variational}
K.-L. Lim, X.~Jiang, Variational posterior approximation using stochastic
  gradient ascent with adaptive stepsize, Pattern Recognition 112 (2021)
  107783.

\bibitem{kohl2018probabilistic}
S.~Kohl, B.~Romera-Paredes, C.~Meyer, J.~De~Fauw, J.~R. Ledsam, K.~Maier-Hein,
  S.~A. Eslami, D.~J. Rezende, O.~Ronneberger, A probabilistic u-net for
  segmentation of ambiguous images, in: Neural Information Processing Systems,
  2018, pp. 6965--6975.

\bibitem{zhang2019variational}
J.~Zhang, C.~Zhao, B.~Ni, M.~Xu, X.~Yang, Variational few-shot learning, in:
  International Conference on Computer Vision, 2019, pp. 1685--1694.

\bibitem{wang2021variational}
H.~Wang, Y.~Yang, X.~Cao, X.~Zhen, C.~Snoek, L.~Shao, Variational prototype
  inference for few-shot semantic segmentation, in: Winter Conference on
  Applications of Computer Vision, 2021, pp. 525--534.

\bibitem{bhat2020learning}
G.~Bhat, F.~J. Lawin, M.~Danelljan, A.~Robinson, M.~Felsberg, L.~Van~Gool,
  R.~Timofte, Learning what to learn for video object segmentation, in:
  European Conference on Computer Vision, 2020, pp. 777--794.

\bibitem{ronneberger2015u_unet}
O.~Ronneberger, P.~Fischer, T.~Brox, U-net: Convolutional networks for
  biomedical image segmentation, in: International Conference on Medical Image
  Computing and Computer Assisted Intervention, 2015, pp. 234--241.

\bibitem{liu2020crnet}
W.~Liu, C.~Zhang, G.~Lin, F.~Liu, Crnet: Cross-reference networks for few-shot
  segmentation, in: Computer Vision and Pattern Recognition, 2020, pp.
  4165--4173.

\bibitem{nguyen2019feature_boost}
K.~Nguyen, S.~Todorovic, Feature weighting and boosting for few-shot
  segmentation, in: International Conference on Computer Vision, 2019, pp.
  622--631.

\bibitem{wang2019panet_panet}
K.~Wang, J.~H. Liew, Y.~Zou, D.~Zhou, J.~Feng, Panet: Few-shot image semantic
  segmentation with prototype alignment, in: International Conference on
  Computer Vision, 2019, pp. 9197--9206.

\bibitem{hu2019attention_a-mcg}
T.~Hu, P.~Yang, C.~Zhang, G.~Yu, Y.~Mu, C.~G.~M. Snoek, Attention-based
  multi-context guiding for few-shot semantic segmentation, in: AAAI Conference
  on Artificial Intelligence, 2019, pp. 8441--8448.

\bibitem{wei2019fss_fss1000}
T.~Wei, X.~Li, Y.~P. Chen, Y.-W. Tai, C.-K. Tang, Fss-1000: A 1000-class
  dataset for few-shot segmentation, in: Computer Vision and Pattern
  Recognition, 2020.

\bibitem{armato2011lung}
S.~G. Armato~III, G.~McLennan, L.~Bidaut, M.~F. McNitt-Gray, C.~R. Meyer, A.~P.
  Reeves, B.~Zhao, D.~R. Aberle, C.~I. Henschke, E.~A. Hoffman, et~al., The
  lung image database consortium (lidc) and image database resource initiative
  (idri): a completed reference database of lung nodules on ct scans, Medical
  physics 38~(2) (2011) 915--931.

\bibitem{zhang2019pyramid}
C.~Zhang, G.~Lin, F.~Liu, J.~Guo, Q.~Wu, R.~Yao, Pyramid graph networks with
  connection attentions for region-based one-shot semantic segmentation, in:
  Computer Vision and Pattern Recognition, 2019, pp. 9587--9595.

\end{thebibliography}


\end{document}